\documentclass[runningheads]{llncs}
\usepackage{graphicx}
\usepackage{tikz}
\usepackage{array}
\usepackage{float}
\usetikzlibrary{positioning}

\tikzset{%
  every neuron/.style={
    circle,
    draw,
    minimum size=1cm
  },
  neuron missing/.style={
    draw=none, 
    scale=4,
    text height=0.333cm,
    execute at begin node=\color{black}$\vdots$
  },
}

\begin{document}
\title{Leveraging WordNet for weight initialization in Neural Language Models}
\author{Ameet Deshpande \and
Vedant Somani}
\institute{Indian Institute of Technology, Madras
% \email{\{abc,lncs\}@uni-heidelberg.de}\\
}
\maketitle              % typeset the header of the contribution
\begin{abstract}
Semantic  Similarity  is  an  important  application  which  finds  it’s  use  in  many  downstream  NLP applications.  Though the task is mathematically defined, semantic similarity’s essence is to capture the  notions  of  similarity  impregnated  in  humans.   Machines  use  some  heuristics  to  calculate  the similarity between words, but these are typically corpus dependent or are useful for specific domains.The difference between Semantic Similarity and Semantic Relatedness motivates the development of new algorithms. \newline For  a  human,  the  word  “car”  and  “road”  are  probably  as  related  as  “car”  and  “bus”.   But this may not be the case for computational methods.  Ontological methods are good at encoding Semantic Similarity and Vector Space models are better at encoding Semantic Relatedness.  There is a dearth of methods which leverage ontologies to create better vector representations. \newline The  aim  of  this  proposal  is  to  explore  in  the  direction  of  a  hybrid  method  which  combines statistical/vector space methods like Word2Vec and Ontological methods like WordNet to leverage the advantages provided by both.
\end{abstract}
\section{Weight Initialization}
\subsection{Motivation}

Pre-training has been recognized for long to be useful to train Neural Networks. The large number of local optimas, and the combinatorial number of equally optimal solutions mean that the initial weight has a large effect on the final answer. Just to represent the combinatorics involved in the problem, consider the following simple Neural Network with just one optimal solution. Fully Connected layers mean that there can be $n!$ different combinations for the same values of weights, which mean that there could be $>n!$ solutions which yield the same globally optimal value. The situation can easily get more complicated for local optimas, something Neural Networks end up being stuck at quite often.

\vspace{1em}

\begin{tikzpicture}[x=1.5cm, y=1.5cm, >=stealth]

\foreach \m/\l [count=\y] in {1,2,3,missing,4}
  \node [every neuron/.try, neuron \m/.try] (input-\m) at (0,2.5-\y) {};

\foreach \m [count=\y] in {1,missing,2}
  \node [every neuron/.try, neuron \m/.try ] (hidden-\m) at (2,2-\y*1.25) {};

\foreach \m [count=\y] in {1,missing,2}
  \node [every neuron/.try, neuron \m/.try ] (output-\m) at (4,1.5-\y) {};

\foreach \l [count=\i] in {1,2,3,n}
  \draw [<-] (input-\i) -- ++(-1,0)
    node [above, midway] {$I_\l$};

\foreach \l [count=\i] in {1,n}
  \node [above] at (hidden-\i.north) {$H_\l$};

\foreach \l [count=\i] in {1,n}
  \draw [->] (output-\i) -- ++(1,0)
    node [above, midway] {$O_\l$};

\foreach \i in {1,...,4}
  \foreach \j in {1,...,2}
    \draw [->] (input-\i) -- (hidden-\j);

\foreach \i in {1,...,2}
  \foreach \j in {1,...,2}
    \draw [->] (hidden-\i) -- (output-\j);

\foreach \l [count=\x from 0] in {Input, Hidden, Ouput}
  \node [align=center, above] at (\x*2,2) {\l \\ layer};

\end{tikzpicture}

Work on Unsupervised Pre-training \cite{unsupervised} showed the immense importance of Weight Initialization. An unsupervised objective is used in their work and they claim that the weight initialization determines which optima the model reaches, as illustrated by the following figure. Staring at $t=0$ and $t=4$ will yield different minimas.

\begin{figure}
\includegraphics[width=\textwidth]{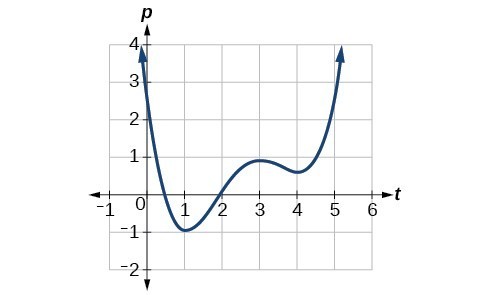}
\caption{Multiple Minimas} \label{fig1}
\end{figure}

Transfer Learning \cite{transfer1} is yet another way to utilize weight initialization. The concept of Transfer Learning has been examined by \cite{transfer2}. The two key constraints are pointed out by them are, what all weights to transfer (relating to co-adaptibility of the layers) and the negative effect on the optimization of the higher level layers (how to build higher level features using lower level features of non-target task). It should be clear that, the fact that we are using a single layer Neural Network (Word2vec) alleviates both these issues.

Our work is similar to Transfer Learning in many aspects, but the fact that the neural networks being used are shallow gives a feel of Weight Initialization rather than reusability of weights. Nevertheless, we show in the experiments section that our method can give better performance than without initialization.

The final motivation to use weight initialization is to reduce the training time. We hypothesize that if the initialized weights are already good enough for most aspects, the number of epochs taken to fine tune them will be much lesser than what would be required to train them from scratch.

% Mention about how one corpus can be used to initialize

\subsection{Using WordNet for initialization}
The importance of weight initialization also makes it crucial to ensure the weights initialized are useful for a large range of tasks. The Word2vec model (detailed in the next section) works with context words. The following characteristics were identified as important to ensure good initialization.

\begin{itemize}
    \item Small Corpus so that training time is reduced
    \item Large vocabulary size to ensure sufficient coverage
    \item Representative Context words even with small corpus
\end{itemize}

A dictionary seemed like a good option which has all the above characteristics. WordNet \cite{wordnet} in particular is arranged in a hierarchy and we thought that the context words and examples used to create them will also encode this hierarchy in some way. Other successes \cite{lesk} of algorithms which use WordNet glosses motivated us to use this corpus.

% Argue why WordNet is a good corpus to initialize with

\section{Method}

The work in \cite{word2vec} introduces models to learn word embeddings in a corpus. Specifically the work introduces the CBOW (Continuous bag of words) model in which the model predicts the word given the context of words. It also introduces the Skipgram architecture which weighs nearby context words more than the distant ones. \newline \newline
% Tell the methodology used. How the Wordnet corpus was generated. For example, we added the word only once, we added it twice etc.
As stated in the earlier section, a dictionary type of corpus was deemed fit for a good initialization. We were motivated to use the WordNet glosses for the same. The model was trained on the word definitions from wordnet to learn the initial word weights. We tried the learning algorithm on different variants of the wordnet gloss corpus. The details of which are as follows : \begin{itemize}
    \item The corpus was created by appending to each word, it's definition. For example "enamel any smooth glossy coating that resembles ceramic glaze" is a part of the corpus and this sentence was formed by concatenating "enamel" and it's gloss "any smooth glossy coating that resembles ceramic glaze". Let us call this \textit{wordnetOnce}.
    \item Another corpus was created by inserting the word into the gloss definition. For example, "enamel any enamel smooth enamel glossy enamel coating enamel that enamel resembles enamel ceramic enamel glaze" is a part of the corpus and this sentence was formed by inserting "enamel" in between it's gloss definition which is "any smooth glossy coating that resembles ceramic glaze". Let us call this \textit{wordnetMultiple}
\end{itemize} 
The model was trained as CBOW and it turned out that the corpus \textit{wordnetOnce} performed better. The similarity between "banana" and "fruit" was reported as 0.442 by wordnetOnce and 0.253 by wornetMultiple. Thus we have used the wordnetOnce corpus for rest of the experiments. \newline
\begin{table}[h]
    \centering
    \begin{tabular}{|c|c|c|}
        \hline
        wordnetOnce & wordnetMultiple& wordnetMultiple \\
        &window size -2 & window size -8   \\
        \hline
        Musa & Monstera & banana \\
        bananas & bananas & Citron \\
        fruit & Treelike & Monstera \\
        Phillippine & perfumed & Glycerine \\
        Hazel & banana & Tent \\
        Shrubby & Crescent-shaped & Apricot \\
        Citrus & Marang & Write \\
        Liquidambar & Anaras & Corozo \\
        buckthorn & Kernels & One-seeded \\ \hline

    \end{tabular}
    \caption{Similar words to banana}
    \label{tab:my_label}
\end{table}
We also experimented if CBOW or Skipgram performs better. We found (subsequent sections) that both the models' performance is very similar and hence we chose to run the experiments using the CBOW model.
% Tell if CBOW was used or SkipGram was used

\section{Evaluation}

\subsection{WordSim}
% Mention about WordSim and the correlation scores
The WordSim-353\cite{wordnetwordsim} dataset contains English word pairs along with human-assigned similarity judgements.During evaluation, we calculate the similarity between the 353 word pairs mentioned in the WordSim-353 dataset  and try to find the correlation between the similarity values as depicted in the dataset compared to the values given by the models. \newline \newline
We use the Spearman correlation metric as opposed to Pearson. Spearman is a correlation test which assesses how well the relationship between two ranked variables is. \newline \newline
\textbf{Spearman:} Spearman is a rank correlation measure which is used to measure the degree of association between the two variables. The following formula is used to calculate Spearman rank correlation: \newline
\begin{figure}[h]
    \centering
    \includegraphics[scale=1]{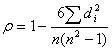}
    \caption{Spearman Rank Correlation}
    \label{spearman}
\end{figure}
\newline
where $d_i$ is the difference between the ranks of corresponding variables and $n$ is the number of observations.

\subsection{Word Analogy}
One of the goals was to test the obtained vectors' performance on the analogy task. The work in \cite{word2vec-analogy} states that the vector representations learnt by the word2vec models capture complex relations like word analogies as well. An analogy task would be of the type - ``What is the word that is similar to \textit{small} in the same sense as \textit{biggest} is similar to \textit{big}?" \newline \newline
The test was run on the family test set present in \cite{analogytestset}. Suppose the test was on \textit{boy:girl :: brother:sister}, we then find the vector $v* = v(brother)-v(sister)+v(boy)$, and then we search for the $10$ closest (via cosine similarity) words to $v*$. We say that the model performed correct for this testcase, if \textit{girl} was among the $10$ most similar words, and we say that the model performed incorrect for the testcase. \newline \newline 
Eventually we report an accuracy measure over $506$ test cases where the accuracy is \[\frac{\#correct\_test\_cases}{\#correct\_test\_cases + \#incorrect\_test\_cases}\] \newline 
The table \ref{analogy} are few sets of words of the form \textit{word1:word2 :: word3:word4} from the test set.
\begin{table}[]
    \centering
    \caption{Analogy task example set}
    \begin{tabular}{|c|c||c|c|}
    \hline
    \textbf{word1} &
    \textbf{word2} &
    \textbf{word3} &
    \textbf{word4} \\ \hline
    boy & girl & brother & sister  \\ \hline
    brother & sister & dad & mom \\ \hline
    father & mother & king & queen \\ \hline
    grandfather & grandmother & grandpa &grandma \\ \hline
    groom & bride & prince & princess \\ \hline
    uncle & aunt & man & woman \\ \hline
    son & daughter & nephew & niece \\ \hline
    \end{tabular}
    \label{analogy}
\end{table}{}

\begin{table}[H]
    \caption{Classifying the performance of model over test-case}
    \centering
    \begin{tabular}{|c|c||c|>{\centering\arraybackslash}m{5cm}|c|c|}
    \hline
    \textbf{word1} &
    \textbf{word2} &
    \textbf{word3} &
    \textbf{10 closest words predicted} &
    \textbf{word4} &
    \textbf{Remark} \\ \hline 
    
    boy & girl & brother & ~\newline brother \newline daughter \newline wife \newline wife, \newline \textbf{sister} \newline son \newline father \newline mother \newline lover \newline nephew \newline &sister & Correct \\ \hline
    
    boy & girl & dad & ~\newline girl \newline chipotle \newline mammal \newline blue-violet \newline thyrsus \newline rosette \newline hedgerow \newline volva \newline lubrica \newline scherzerianum \newline & mom & Incorrect \\ \hline
    \end{tabular}
    \label{tab:my_label}
\end{table}
\section{Experiments}

\subsection{Corpus Details}
For the experiments conducted we primarily used the British National Corpus\cite{bnc} and the partial wikipedia corpus\cite{wikiCorpus}. In particular, we will refer to the following corpora as mentioned below.
\begin{itemize}
    \item B - This is subset `B' of the British National Corpus
    \item AB - This is the concatenation of `A' and `B' subsets.
    \item ABC - This is the concatenation of `A', `B' and `C' subsets
    \item Partial Wikipedia / enWiki - This dataset comprises of the first billion characters from wikipedia. This amounts to less than 10\% of the information available on the wikipedia and can be found in \cite{wikiCorpus}
\end{itemize}

\subsection{Skipgram vs CBOW}

We ran the experiments on the AB corpus once using the skipgram model and then using the CBOW model, both with a window size of 8. The observations are plotted below:
\begin{figure}[H]
    \centering
    \begin{minipage}{0.4\textwidth}
    \centering
    \includegraphics[width=\textwidth]{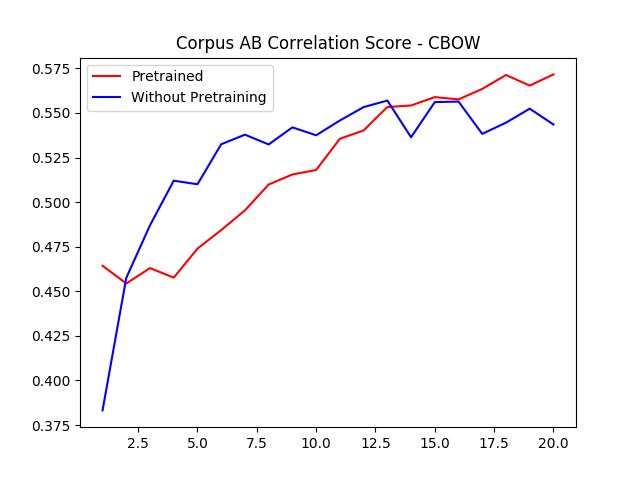}
    \caption{CBOW}
    \label{CBOW}
    \end{minipage}
    \begin{minipage}{0.4\textwidth}
    \centering
    \includegraphics[width=\textwidth]{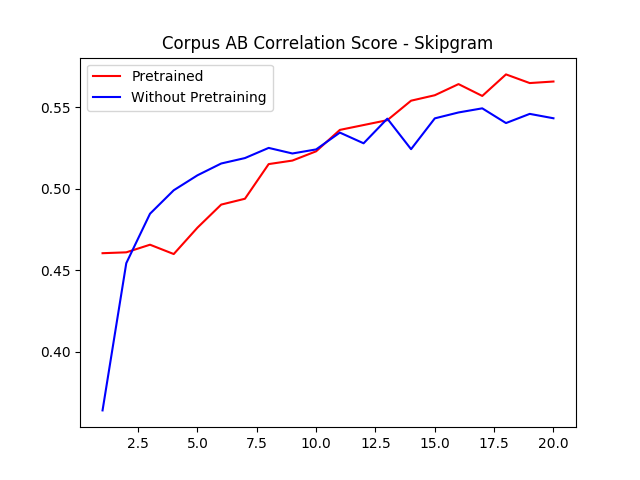}
    \caption{Skipram}
    \label{Skipgram}
    \end{minipage}
    \centering
    \begin{minipage}{0.4\textwidth}
    \centering
    \includegraphics[width=\textwidth]{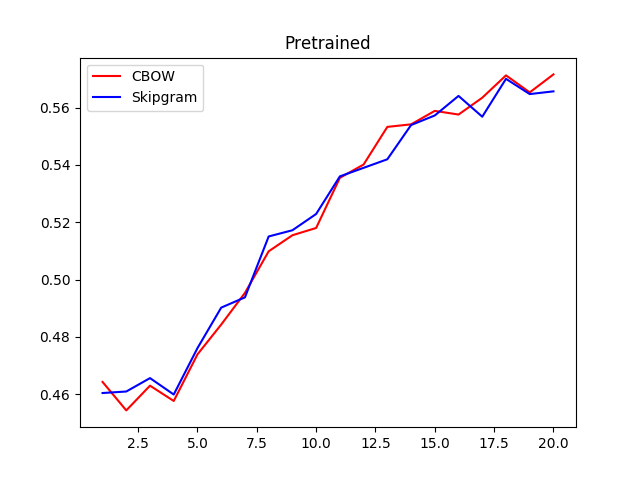}
    \caption{Pretrained}
    \label{cbowskipgram}
    \end{minipage}
    \caption{CBOW vs Skipgram}
    \label{cbowVsSkipgram}
\end{figure}
We conclude that the observations are very close, and there is no significant difference in performance. Hence we chose to run the further experiments on the CBOW model.

\subsection{Performance of pretrained vectors compared to non-pretrained vectors given equivalent training}

Four corpora, referred as B, AB, ABC and Partial Wikipedia were used to run these experiments. Vector embeddings were learnt by the CBOW word2vec model with a window size of 8. \newline \newline
Let us call, the word vectors learnt on the \textit{wordnetOnce} corpus as \textit{wordnetVectors}. The experiments were run once by initializing the word vectors to wordnetVectors. We call this the Pretrained setup, and the they were then run without any extra initialization, which we refer to as the Without Pretraining Setup. Following are the observations of how the vectors performed on
correlation scores with the wordsim.

\begin{figure}[H]
    \centering
    \begin{minipage}{0.4\textwidth}
    \centering
    \includegraphics[width=\textwidth]{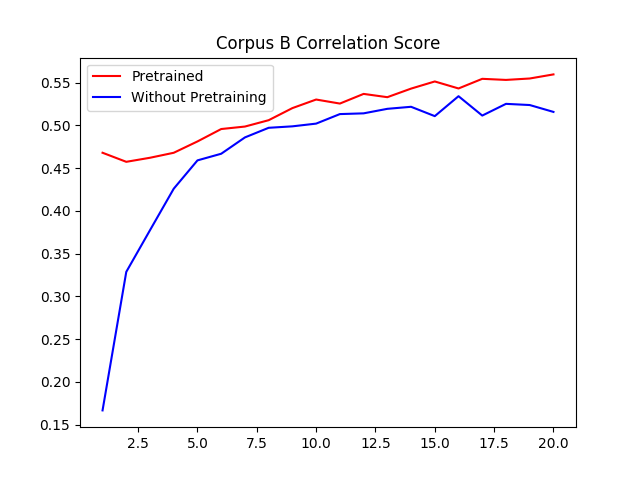}
    \caption{Corpus B}
    \label{B_corr}
    \end{minipage}
    \begin{minipage}{0.4\textwidth}
    \centering
    \includegraphics[width=\textwidth]{AB_CBOW_correlation.png}
    \caption{Corpus AB}
    \label{AB_corr}
    \end{minipage}
    \centering
    \begin{minipage}{0.4\textwidth}
    \centering
    \includegraphics[width=\textwidth]{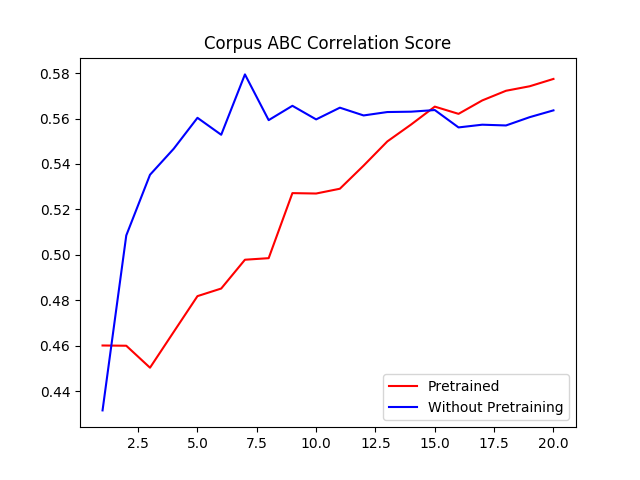}
    \caption{Corpus ABC}
    \label{ABC_corr}
    \end{minipage}
    \begin{minipage}{0.4\textwidth}
    \centering
    \includegraphics[width=\textwidth]{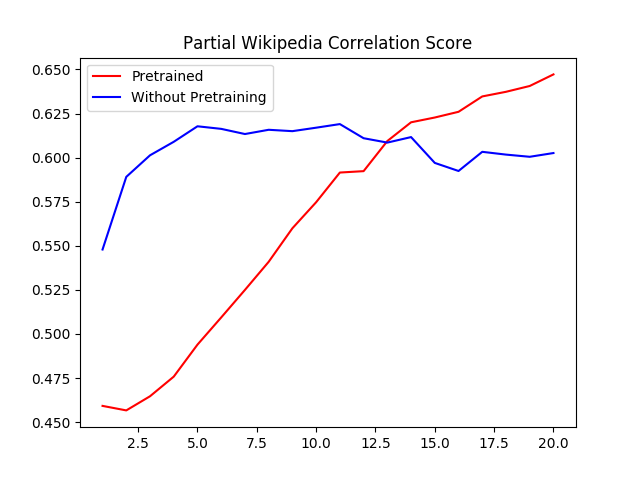}
    \caption{Corpus Partial Wikipedia}
    \label{wiki_corr}
    \end{minipage}
    \caption{Correlation Score - Pretrained vs Without Pretraining}
    \label{corr}
\end{figure} 
From the figure \ref{corr} we can observe that after sufficient amount of training, the experiments in which the vectors were pretrained give better correlation scores with wordsim when compared to the ones without any particular word vector initialization. The effect is more clearly visible when the training for word vectors is done for more number of epochs. The partial wikipedia corpus when trained for \textbf{40 epochs}, gave a correlation score of \textbf{0.6598} when the word vectors were initialized with pretrained word vectors on wordnetOnce corpus as compared to the correlation score of \textbf{0.5759} when there was no vector initialization done. \newline \newline
Given above evidence we conclude that vectors pretrained with wordnetOnce corpus give better similarity scores than the vectors which are learnt without any pretraining. 

\subsection{Given the correlation score to achieve compare training time across corpuses}
The aim of this experiment was to find out the effect of pretraining in the training time. The experiment goal was to find out the number of epochs of pretrained vectors after which the correlation score with wordsim was greater than the correlation score obtained by training the vectors without initialization for 20 epochs. The observations are listed in table \ref{numEpochs}.

\begin{table}[H]
    \centering
    \caption{Performance for desired correlation score}
    
    \begin{tabular}{|c|c|c|c|}
    \hline
    \textbf{Corpus} & \textbf{Size} & \textbf{Correlation Score at 20 epochs} & \textbf{\# epochs for} \\
    & & \textbf{Vectors without pretraining} & \textbf{Pretrained vectors} \\ \hline 
    B & 39MB & 0.5159 & 9  \\ \hline
    AB & 118MB & 0.5435 & 13 \\ \hline
    ABC & 217MB & 0.5636 & 15 \\ \hline
    Partial Wikipedia & 954MB & 0.6026  & 13 \\ \hline
    \end{tabular}
    \label{numEpochs}
\end{table}

From Table \ref{numEpochs} it is quite evident that \textbf{pretraining helps reduce the training time} too, while trying to achieve a particular correlation score.

\subsection{Variation of correlation score for a given training time, varying size of corpus}

Trying to understand the effect of pretraining across corpora of different sizes, we split the partial wikipedia (enwiki) corpus into parts of 239MB, 477MB, 716MB and 954MB (full corpus) and learnt the vector representations in the case of pretraining and also without pretraining. The training algorithm was run for 20 epochs and the observations are reported in table \ref{corpusSize}.

\begin{table}
    \centering
    \caption{Effect on varying corpus size}
    \begin{tabular}{|c|c|c|c|}
    \hline
    Corpus & Size & Correlation Score Without Pretraining & Correlation Score With Pretraining \\ \hline 
    $\frac{enwiki}{4}$ & 239MB & 0.6113 & 0.6479 \\ \hline
    $\frac{2*enwiki}{4}$ & 477MB & 0.6215 & 0.6565 \\ \hline
    $\frac{3*enwiki}{4}$ & 716MB & 0.6186 & 0.6479 \\ \hline
    $\frac{4*enwiki}{4}$ & 954MB & 0.6026 & 0.6472 \\ \hline
    \end{tabular}
    \label{corpusSize}
\end{table}

One of the observations we can draw is that after 20 epochs all the above corpora perform better when they are pretrained than when no pretraining is done. Another interesting observation is that there is no significant difference in correlation score across different sizes, owing to the similar nature of the corpora. Thus for better vector representations we do not necessarily have to train with full data.
\subsection{Performance on analogy task}
\begin{figure}[H]
    \centering
    \begin{minipage}{0.4\textwidth}
    \centering
    \includegraphics[width=\textwidth]{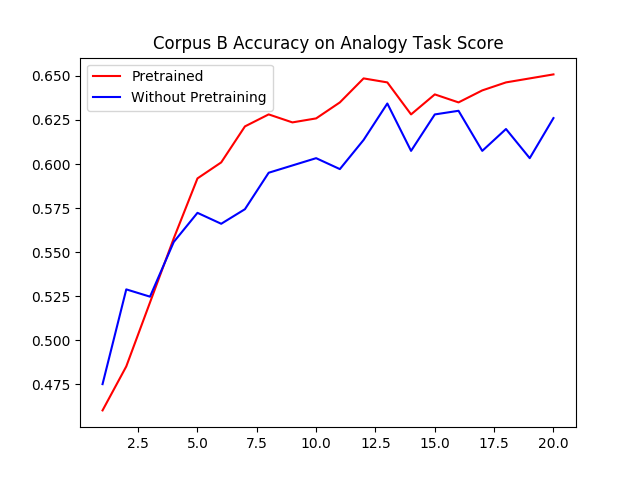}
    \caption{Corpus B}
    \label{B_accu}
    \end{minipage}
    \begin{minipage}{0.4\textwidth}
    \centering
    \includegraphics[width=\textwidth]{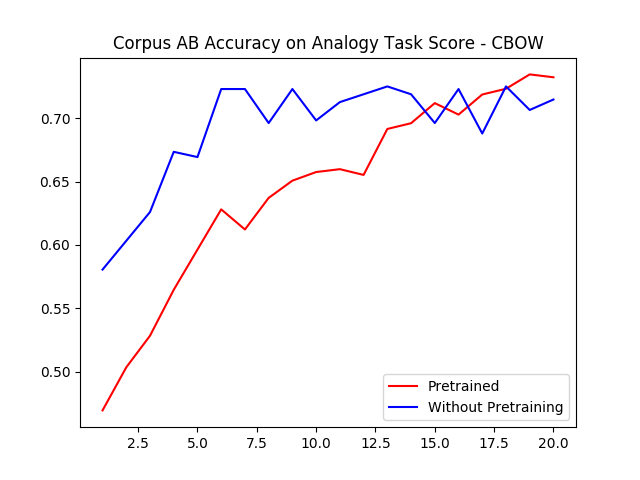}
    \caption{Corpus AB}
    \label{AB_accu}
    \end{minipage}
    \centering
    \begin{minipage}{0.4\textwidth}
    \centering
    \includegraphics[width=\textwidth]{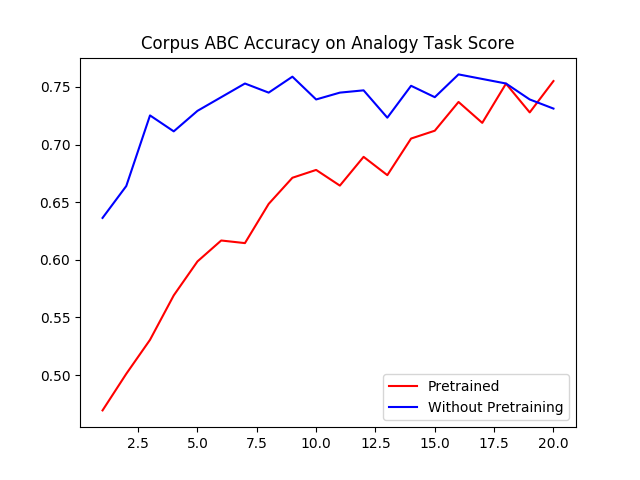}
    \caption{Corpus ABC}
    \label{ABC_accu}
    \end{minipage}
    \begin{minipage}{0.4\textwidth}
    \centering
    \includegraphics[width=\textwidth]{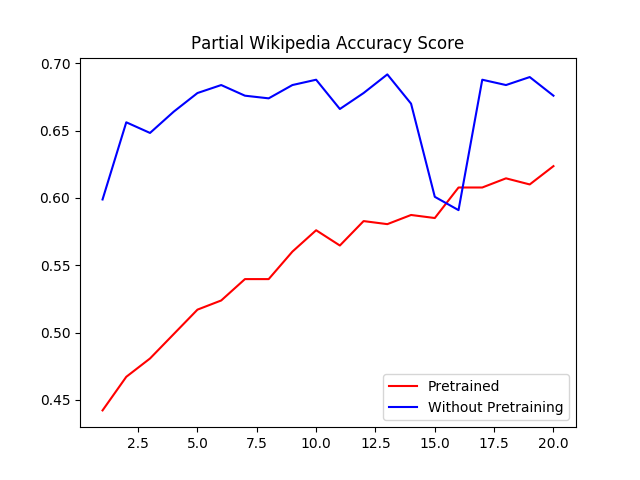}
    \caption{Corpus Partial Wikipedia}
    \label{wiki_accu}
    \end{minipage}
    \caption{Accuracy Score on Analogy Task - Pretrained vs Without Pretraining}
    \label{accu}
\end{figure}

For the experiments conducted, evaluation was also done on the accuracy score on the analogy task as described in the earlier sections. The results are plotted in the figure \ref{accu}.

On this made up Word Analogy task as well, we can see that Pre-training seems to giving a better analogy score than without pre-training. On the partial Wikipedia corpus however, we don't see a better score. These experiments are not as decisive as the previous experiments, but we could still claim that it is giving an improvement on smaller corpus sizes. The Partial Wikipedia corpus is slightly bigger than the other $3$ corpuses considered.\newline \newline
We thus conclude that in the best case pre-training helps in Analogy tasks for small corpuses and in the worst case it does not degrade the performance. Note the Word Analogy is almost never the main property we look for in Word Vectors. Correlation score or some extrinsic measure is more reliable because of the nuances present in Word Analogy.

% We can observe that we cannot clearly say that pretraining helped in increasing the accuracy as we can see in Figure \ref{wiki_accu} that the performance was better in the case of non-pretrained case. But then in the rest of the cases, it seems to have helped. \newline \newline 
% These results do not allow us to make conclusions with high confidence and we thus refrain from the same. But however we note that 

\section{Domain Transfer}
% This part will house the WordNet to sci.med transfer
Since WordNet is a general corpus, it may not be able to model domain specific vectors well. It is well known that ``meaning" changes with the domain. Domain adaptation aims to minimize the computation so as to have vectors which are suited to multiple domains. Experiments were conducted to check the transferability of the WordNet vectors to a relatively niche domain. More specifically, the \textit{sci.med} category from 20 Newsgroups \cite{20} was used to see how well WordNet vectors can adapt to the Medical Domain.

Experiments are conducted with and without pre-trained vectors and evaluation is four fold.

\begin{enumerate}
    \item Correlation scores on WordSim
    \item Word Analogy score
    \item Similarity of words from medical domain
    \item Similarity of general words
\end{enumerate}

The motivation for the first three parts is clear and the fourth evaluation is to check if training on another corpus degrades the previously encoded useful information. We definitely expect some decrease in the same, but the disparity between the values is what we want to check.\\

The following are the set of words that are used for the medical domain. The words were chosen such that each word has at least $25$ occurrences in \textit{sci.med} corpus, and a few of the pairs had sufficiently different meaning in the medical domain as compared to colloquial language.

\begin{table}[H]
\caption{Medical Words}\label{medical}
\begin{center}
    \begin{tabular}{||c|c||}
    \hline
doctor& nurse\\ \hline
doctor& syringe\\ \hline
doctor& medicine\\ \hline
syringe& medicine\\ \hline
hospital& nurse\\ \hline
disease& medicine\\ \hline
hospital& nurse\\ \hline
hospital& health\\ \hline
health& medicine\\ \hline
hospital& problem\\ \hline
treatment& cancer\\ \hline
breast& cancer\\ \hline
database& medical\\ \hline
depression& medicine\\ \hline
depression& chemical\\ \hline
family& planning\\ \hline
children& vaccine\\ \hline
    \end{tabular}
\end{center}
\end{table}

Similarly, the following were the general words used.

\begin{table}[H]
\caption{General Words}\label{general}
 \begin{center}
    \begin{tabular}{||c|c||}
    \hline
school& children\\ \hline
university& students\\ \hline
company& industry\\ \hline
boy& girl\\ \hline
mother& father\\ \hline
national& international\\ \hline
library& books\\ \hline 
    \end{tabular}
    \end{center}
\end{table}

The words from Table \ref{medical} are expected to have a high similarity when trained only on the \textit{sci.med} corpus because of high occurrence and less polysemy in domain specific corpus. The following are the similarity scores and their comparisons. The training was done for $20$ epochs.

\begin{table}[H]
\caption{Comparisons on words from Medical Domain}\label{medical_dom}
\begin{center}
    \begin{tabular}{||c|c|c|c|c||}
    \hline
    Word 1 & Word 2 & Sci.med & Pretrained & WordNet\\ \hline
doctor& nurse& 0.479  & 0.614 & 0.672 \\ \hline
doctor& syringe& 0.180 & 0.120 & 0.156\\ \hline
doctor& medicine& 0.321 & 0.366 & 0.432\\ \hline
syringe& medicine& 0.237 & 0.208 & 0.329\\ \hline
hospital& nurse& 0.281 & 0.449 & 0.537\\ \hline
disease& medicine& 0.255 & 0.357 & 0.438\\ \hline
hospital& nurse& 0.281 & 0.449 & 0.537\\ \hline
hospital& health& 0.390 & 0.459 & 0.518\\ \hline
health& medicine& 0.353 & 0.415 & 0.492\\ \hline
hospital& problem& 0.258 & 0.274 & 0.268\\ \hline
treatment& cancer& 0.564 & 0.555 & 0.652\\ \hline
breast& cancer& 0.647 & 0.508 & 0.588\\ \hline
database& medical& 0.344 & 0.413 & 0.485\\ \hline
depression& medicine& 0.200 & 0.162 & 0.227\\ \hline
depression& chemical& 0.352 & 0.229 & 0.256\\ \hline
family& planning& 0.326 & 0.256 & 0.231 \\ \hline
children& vaccine& 0.579 & 0.233 & 0.151 \\ \hline
    \end{tabular}
\end{center}
\end{table}

The conclusions that can be drawn from the above table are, for words on which WordNet already has a high similarity, the pre-trained words maintain that similarity, and for word pairs like ``family, planning" and ``children, vaccine" which have a low similarity in WordNet and higher similarity in \textit{sci.med}, the similarity of pre-trained vectors increases.

The results in table \ref{general_dom} are for the general words. As expected, the pretrained vectors perform way better.

\begin{table}[H]
\caption{Comparisons on general words}\label{general_dom}
 \begin{center}
    \begin{tabular}{||c|c|c|c|c||}
    \hline
    Word 1 & Word 2 & Sci.med & Pretrained & WordNet \\ \hline
school& children & 0.034 & 0.462 & 0.463\\ \hline
university& students& 0.275 & 0.425 & 0.435\\ \hline
company& industry & 0.0434 & 0.524 & 0.520\\ \hline
boy& girl& 0.089 & 0.804 & 0.863\\ \hline
mother& father& 0.613 & 0.829 & 0.849\\ \hline
national& international& 0.687 & 0.650 & 0.687\\ \hline
library& books& 0.435 & 0.566 & 0.560\\ \hline 
    \end{tabular}
    \end{center}
\end{table}

\begin{figure}[H]
\includegraphics[width=\textwidth]{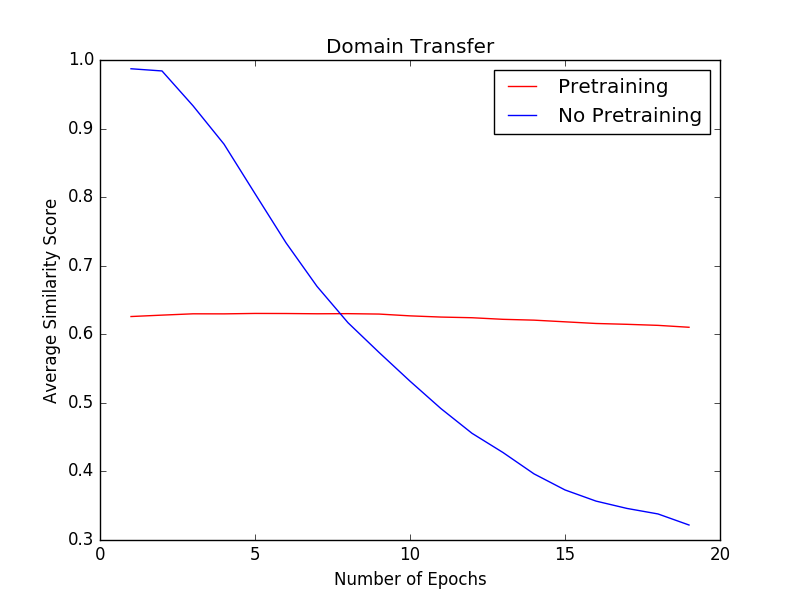}
\caption{General Words} \label{all_words_plot}
\end{figure}

From the results in table \ref{general_dom}, it can be seen that the pre-trained vectors are able to capture the nuances of the medical domain, while largely maintaining the information it obtained when trained only on WordNet. Though there is some loss in similarity with respect to the vectors trained on just WordNet, the loss is very small in \textbf{all} the cases. We can conclude that WordNet can be adapted well in more niche domains because it naturally augments the knowledge it has acquired.
\newline
The words we have chosen to measure similarity are such that we want their correlation to be as high as possible. To pictorially represent the results, the following procedure was followed. For both set of words above, we measure the similarity of all the words after each epoch and then take an average. Ideally, the average should be taken, but since we created our own set weighing the word pairs based only on our prior experience may bias the results. We stuck to uniform weight for each word. Figure \ref{all_words_plot} plots the results for words in table \ref{general}.

\begin{figure}[H]
\includegraphics[width=\textwidth]{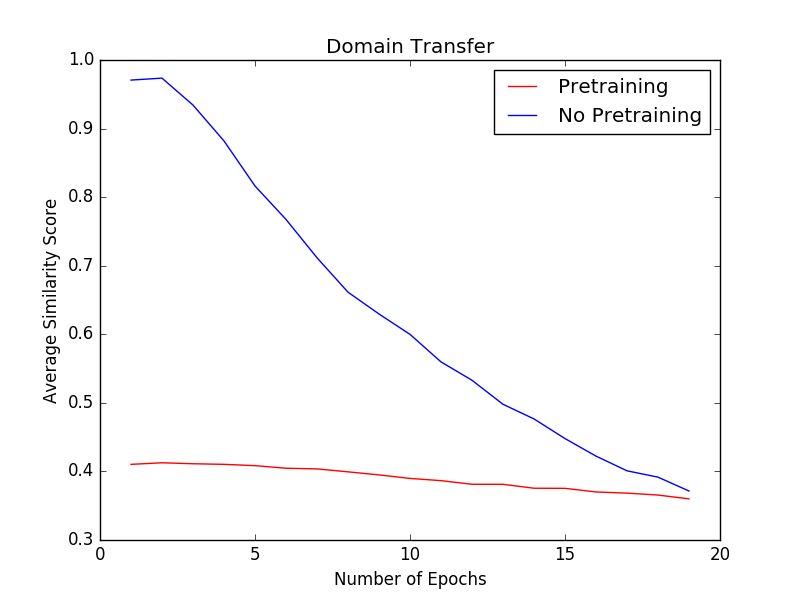}
\caption{Domain Specific Words} \label{words_plot}
\end{figure}

As expected, the average score stays almost the same for pre-trained vectors because it has already been trained to do well on this data. Without pre-training, \textit{sci.med} corpus is not able to find the correlation. We perform the same experiment with the domain specific words and the plot it in Figure \ref{words_plot}.

Figure \ref{words_plot} shows that the score for \textit{Pretraining} remains almost constant, not changing much. The score for without Pre-training is that high initially because of the fact that all the similarities start from $0$. Though the similarity score remained constant we could see that the scores are changing for many values. So we decided to measure the variance in scores to check if there are any changes.

\begin{figure}[H]
\includegraphics[width=\textwidth]{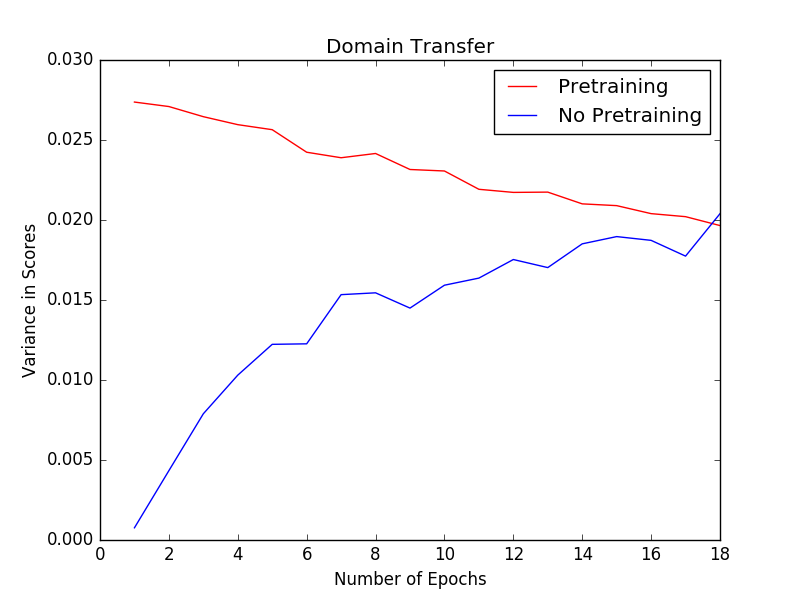}
\caption{Variance of scores} \label{variance_domain}
\end{figure}

Results in Figure \ref{variance_domain} are pretty revealing. Though the sum of the scores are remaining constant, the similarity is being distributed among words. This is because some words in the WordNet corpus have very high similarity as compared to the domain corpus and some words had lesser correlation. After seeing the new corpus, the model is trying to even out the scores because it has ``realised" (through data) that the words are all somewhat equally similar, which is pretty amusing!

\section{Conclusion}
We observed the effect of weight initialization over the word vectors and correlation with the wordsim353 similarity scores. We conclude that when the weight initialization with word embeddings learnt over a dictionary like corpus, the pretrained vectors perform better at word similarity task than the non-pretrained vectors. To reach an equivalent correlation score, the pretrained vectors need lesser training time. In one of the experiments we also showed that we do not need the full corpus. A good representation subset of a corpus also gives similar performance, although a less sized corpus implies lesser training time.

\end{document}